\documentclass{ws-procs11x85}
\usepackage{ws-procs-thm}           
\usepackage{graphicx}
\begin{document}

\title{From Promising Capability to Pervasive Bias: Assessing Large Language Models for Emergency Department Triage
}





\author{Joseph Lee\textsuperscript{\rm 1}, Tianqi Shang\textsuperscript{\rm 1}, Jae Young Baik\textsuperscript{\rm 1}, Duy Duong-Tran\textsuperscript{\rm 1}, \\ Shu Yang\textsuperscript{\rm 1,*}, Lingyao Li\textsuperscript{\rm 2,*}, Li Shen\textsuperscript{\rm 1,*}}

\address{\textsuperscript{\rm 1}University of Pennsylvania,
Philadelphia, 19104, USA\\
\textsuperscript{\rm 2}University of South Florida,
Tampa, 33620, USA\\
E-mail: jojolee@seas.upenn.edu, tianqi.shang@pennmedicine.upenn.edu, 
duyanh.duong-tran@pennmedicine.upenn.edu, jaybaik@sas.upenn.edu,
shu.yang@pennmedicine.upenn.edu, 
lingyaol@usf.edu,
li.shen@pennmedicine.upenn.edu\\
$*$: Correspondence Authors}

\begin{abstract}
Large Language Models (LLMs) have shown significant promise in clinical decision support, yet their application to triage remains underexplored. In this study, we systematically investigate the capabilities of LLMs in emergency department triage through two key dimensions: (1) robustness to distribution shifts and missing data, and (2) counterfactual analysis of intersectional biases across sex and race. We assess multiple LLM-based approaches, ranging from continued pre-training to in-context learning, as well as conventional machine learning approaches. Our results indicate that LLMs exhibit superior robustness, and we further investigate the key factors contributing to the promising LLM-based approaches. Moreover, in this setting, we also identify critical gaps in LLM preferences that emerge in particular at the intersections of sex and race. LLMs generally exhibit sex-based differences, but they are most pronounced in certain racial groups. These findings suggest that LLMs, while offering powerful capability for clinical triage, encode demographic preferences that may emerge in specific clinical contexts or particular combinations of characteristics, which demand rigorous auditing before real-world integration.

\end{abstract}

\keywords{Emergency Department Triage; Large Language Models; Artificial Intelligence; Evaluation; Clinical Informatics.}

\copyrightinfo{\copyright\ 2024 The Authors. Open Access chapter published by World Scientific Publishing Company and distributed under the terms of the Creative Commons Attribution Non-Commercial (CC BY-NC) 4.0 License.}

\section{Introduction}\label{aba:sec1}
Recent advancements of Large language models (LLMs) have shown promise in various clinical settings \cite{cascella2023evaluating,park2024assessing, li2024scoping, jiang2023health}, with example applications ranging from clinical decision support \cite{liu2023using} to conversational systems \cite{montagna2023data}. Amongst these applications, emergency departments (EDs) are some of the most dynamic and high-stakes clinical environments, serving as the initial point of contact for patients requiring urgent medical attention \cite{franc2024accuracy}. A critical component of ED operations is the triage process, where patients are prioritized based on clinical urgency to ensure timely and appropriate care. Overtriage misallocates scarce resources and undertriage endangers time-sensitive patients \cite{chmielewski2022esi}.  In an era of historic ED overcrowding, these issues are a significant concern \cite{friedman2024artificial}. While triage is an imperfect practice due to limited information \cite{hinson2018accuracy}, triage practices performed by humans are further prone to challenges such as variability in decision-making, lack of training programs, cognitive biases, and decreased quality during overwhelming patient arrivals \cite{joseph2023race, chmielewski2022esi}. 

\begin{figure}[h]
\centerline{
  \includegraphics[width=\textwidth]
  {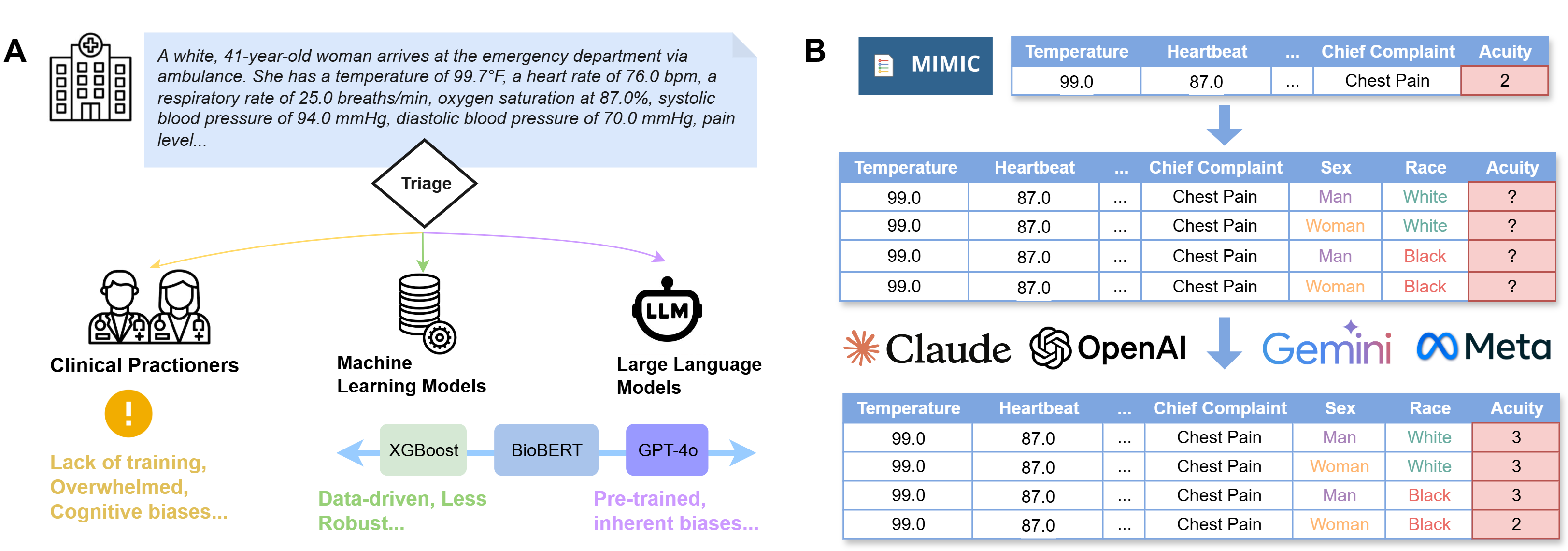} 
}
  \caption{\textbf{A.} represents our evaluation of various models \textbf{B.} denotes our counterfactual analysis, where we perturb the dataset by treating demographics as treatment variables to examine intersectional bias.}
  \label{fig1}
\end{figure}

Traditional machine learning (ML) algorithms have seen success in clinical triage \cite{hong2018predicting, goto2019machine, miles2020using}, but they are limited by their reliance on structured input data and sensitivity to distributional shifts. In practice, much of clinical data consists of unstructured text, such as clinical notes, which can be challenging to utilize in conventional ML. Moreover, substantial variation in triage distributions across hospitals has been documented \cite{chmielewski2022esi}, posing additional challenges for model generalisability\cite{li2024scoping}. 

In contrast, the pre-trained clinical knowledge \cite{singhal2023large} of LLMs and their reasoning abilities pose them as a robust model, able to adapt to diverse settings. Beyond mere prediction, LLMs can generate explanatory rationales, serving as a form of second opinion for clinicians and potentially mitigating mispractice \cite{goh2023chatgpt}. With their potential to improve the efficiency, accuracy, and equity of emergency department triage \cite{preiksaitis2024role}, LLMs are well-positioned to meet the growing demand for AI-driven decision support, an area already seeing early real-world adoption \cite{friedman2024artificial, johns2022insight}.

Despite their potential, the application of LLMs to triage remains an understudied area. Few studies have examined their capabilities \cite{williams2024use, masanneck2024triage, lu2024triageagent}, and they predominantly focus on basic prompting techniques without testing advanced methodologies. In addition, we find that most studies have not replicated realistic triage scenarios, usually limited to a brief description of symptoms, pain, and vital signs, rather than in-depth clinical notes \cite{friedman2024artificial}. 

Furthermore, a concern comes from the potential biases and inconsistencies in LLMs when applied to triage decisions, where unfair prioritization leads to adverse outcomes against certain demographics. While research has documented various forms of bias in LLMs across different healthcare contexts \cite{guevara2024large, omiye2023large, zack2024assessing, yang2024unmasking}, the implications for hospital triage remain unclear. This knowledge gap is critical given that triage can affect vulnerable patient populations. The high-stakes nature of triage decisions necessitates an extensive assessment of LLMs' performance across diverse patient populations and clinical scenarios before any potential deployment.

Addressing these research gaps is important for advancing the integration of LLMs into clinical environments, and achieving this requires stronger collaboration between the medical and natural language processing (NLP) communities. Toward these goals, our paper makes the following key contributions:

\begin{enumerate}
    \item We comprehensively evaluate across distribution shifts, missing data, and datasets ranging from real hospital records to curated case studies. We assess various LLM-based approaches, ranging from continued pre-training to in-context learning, providing insights for future research directions.
    \item We propose a novel counterfactual analysis framework to systematically investigate potential biases in LLM predictions, with particular attention to intersections of sex and race. To the best of our knowledge, we are the first to look at the intersectional bias of LLMs, particularly in the clinical setting.
\end{enumerate}

\section{Related Works}


\textbf{LLMs in Biomedical and Clinical Applications.} 
LLMs have been extensively integrated into the biomedical and clinical domain through a range of approaches, from fine-tuning general LLMs with biomedical-specific corpora~\cite{yunxiang2023chatdoctor, shu2023visual, chen2023meditron, zhang2023biomedgpt, labrak2024biomistral} to equipping frozen LLMs with techniques like chain-of-thought (CoT) reasoning and retrieval-augmented generation (RAG)~\cite{jeong2024improving, toufiq2023harnessing,lee2024knowledgedrivenfeatureselectionengineering,li-etal-2024-dalk, zakka2024almanac, lozano2023clinfo, shang2024leveragingsocialdeterminantshealth}. 
In our study, we focus on a few of these recent approaches that have yet to be applied to triage. For example, studies have shown that careful selection of examples can significantly impact model performance  \cite{ rubin2021learning} and few-shot prompting techniques, such as retrieving k-nearest neighbors (KNN), have been proposed to enhance LLM performance by retrieving relevant examples \cite{xu2023knn}. Furthermore, recent studies have explored how LLMs acquire knowledge and expertise \cite{ovadia2023fine, zhou2023lima} and we extend these investigations to our specific setting.


\textbf{LLMs in the Emergency Department.} Recent studies have explored LLMs in triage tasks. For example, a study evaluated the performance of LLMs on a binary triage task, comparing two patients and determining which presented higher acuity. However, their approach relied on clinical notes generated after the fact, which would be unavailable at the point of triage~\cite{williams2024use}. Another study assessed triage accuracy using CoT prompting and found that conditioning on demographic information did not affect model performance. However, the limited scope of their dataset, 45 vignettes, restricts the generalizability of their findings ~\cite{ito2023accuracy}. Williams extended the study of LLMs to emergency department scenarios such as predicting hospital admission, the need for radiological investigations, and antibiotic prescriptions. The models’ performance remained significantly lower than that of resident physicians across most tasks~\cite{williams2024evaluating}.  The most advanced work to date is an LLM-based agent system that leverages RAG~\cite{lu2024triageagent}. However, the validity of their findings is constrained by their small sample size and inconsistencies we've found in their reported results. In real-world triage, information is limited and frequently incomplete, posing challenges for LLMs operating in dynamic clinical settings. We believe this calls for better benchmarking, which has emerged in other settings \cite{li2024mediq, ouyang-etal-2024-climedbench}. These works represent a crucial step forward, but further research is needed to evaluate the readiness of LLMs.

\textbf{LLM Bias in Clinical Settings.} Recent studies have highlighted significant biases in LLMs, particularly within clinical contexts. For instance, LLMs have been found to exhibit covert rather than overt biases, making such biases less detectable and potentially more harmful\cite{hofmann2024ai}. The study identified biases against individuals speaking African American English dialects, illustrating the subtle ways in which LLMs can perpetuate inequities. Another study found that when asking GPT-4 to describe a case of sarcoidosis, the model produced a vignette of a Black patient 97\% of the time and a Black female patient 81\% of the time \cite{zack2024assessing}. Similarly, Yang employed counterfactual analysis to demonstrate that LLMs' biased behaviors often align with existing healthcare disparities\cite{yang2024unmasking}. For instance, white patients were systematically provided better treatment recommendations than Black patients, despite identical clinical information. However, to date, no research has systematically explored the intersectionality of sex and race in these biases—an important gap that we aim to address in this work.

\section{Methods}
Our study is structured into two parts: (1) a comprehensive evaluation of LLMs on the task of emergency department triage and (2) a counterfactual analysis that inserts demographic attributes into patient profiles. 


\subsection{Evaluating the Clinical Utility of LLMs}

\textbf{Datasets.} We utilize three datasets: (1) hospital records from MIMIC-IV-ED v2.2 \cite{Johnson2023MIMICIVED}, which follow the ESI protocol; (2) hospital records using the Korean Triage Acuity System (KTAS) \cite{moon2019triage}  which were subsequently re-labeled in hindsight by three experts; and (3) curated case studies from the ESI handbook \cite{Gilboy_2005}.

Each dataset includes vital signs (temperature, heart rate, respiratory rate, etc.), self-reported pain levels (1–10), and free-text chief complaints recorded at triage. Acuity is categorized into five levels for both ESI and KTAS, with Level 1 denoting critical cases requiring immediate intervention and Level 5 indicating non-urgent cases.

These datasets differ in two key dimensions. First, these datasets differ in the reliability and nature of their ground truth labels, which affects how we recommend interpreting results based on them. MIMIC labels reflect nurses' initial assessments, which are prone to mistakes, KTAS labels benefit from expert consensus that was formed after the fact, and the ESI handbook comprises of curated case studies. Second, information availability varies, with KTAS containing additional structured features beyond MIMIC, and the ESI handbook providing narratives with comprehensive details.

\textbf{Benchmark Creation.}We construct our benchmark with carefully controlled train-test distributions. For MIMIC-IV-ED, we partition data by temporal cohorts, using 10,000 patients from 2014–2016 for training and 1,000 patients from 2017–2019 for testing. Furthermore, we apply stratified sampling across missingness (and acuity levels) to ensure representation of incomplete records. For KTAS, we incorporate a domain shift by training on 689 patients from a local hospital and testing on 580 patients from a separate regional hospital, though both are academic, urban medical centers. Finally, for the ESI handbook, we adopt the split from Lu\cite{lu2024triageagent}.

\textbf{Baselines.}
We implement two ML baselines (Logistic Regression, XGBoost) to see how traditional methods fare in the task of triage. Additionally, to incorporate the chief complaint (only available as text), we implement a baseline based on BioBERT \cite{lee2020biobert}. Lastly, we implement eight LLM-based approaches: zero-shot with vanilla prompting, zero-shot with CoT prompting \cite{wei2022chain}, zero-shot with automatic CoT (AutoCoT) prompting \cite{zhang2022automatic}, few-shot with vanilla prompting, few-shot with CoT prompting, few-shot using k-nearest neighbors (KNN) retrieval i.e. KATE \cite{liu-etal-2022-makes}, KATE with CoT prompting, and supervised fine-tuning on top of continued pre-training, which is the common recipe to adapt LLMs for a particular domain. Further details on the implementation of our baselines can be found in the later section~\ref{sec:appendix_A}. 

\textbf{Metrics.} Our evaluation framework assesses performance across Accuracy, F-1 Score, Quadratic-Weighted Kappa (QWK), and Mean Square Error (MSE). QWK, a variant of Cohen's Kappa, is used to measure agreement between the model’s predictions and human triage decisions.

\subsection{Counterfactual Analysis of Demographic Bias}
In this section, we assess potential biases present in LLMs in clinical settings. Our approach focuses on counterfactual analysis, enabling a systematic assessment of how intersectional, demographic attributes influence model predictions.

\textbf{Benchmark Dataset.} For this analysis, we leverage the publicly available Demo portion of the MIMIC-IV-ED dataset, as the dataset is accessible to all.

To construct the dataset, we augment each sample in the MIMIC-IV-ED-Demo dataset by inserting altered demographic attributes. Specifically, we generate all 12 counterfactual variations by considering the full combination of two sex categories (Male, Female) and six race categories (White, Black, Asian, Hispanic, American Indian, and Native Hawaiian/Asian Pacific Islander). This expansion reflects the typical granularity of race categories documented in electronic health records (EHRs). We focus on the intersection of demographics—combinations of sex and race—where biases can be amplified.

\textbf{Experimental Setup.} The counterfactual benchmark is evaluated on prominent LLMs: Llama-3.1-70B-Instruct, Gemini-2.0-Flash, gpt-4o-mini, gpt-4o, claude-3-haiku, claude-3-sonnet, and o3-mini (low reasoning). A temperature of 0 is used for all models. To investigate the impact of reasoning strategies on model bias, we further test some of the models with CoT prompting.

\textbf{Statistical Analysis.} We treat sex and race as independent treatment variables and conduct statistical tests to evaluate the significance of their effects. We apply the Wilcoxon signed-rank test to compare model outputs across male and female variations of the same sample. We also apply the Friedman test to evaluate differences in model predictions across the six race categories and their intersections with sex. To account for the testing of multiple LLMs and strategies (e.g. CoT), we apply the Bonferroni correction to control for the false discovery rate in our analysis.

\subsection{Implementation Details}
\label{sec:appendix_A}
For our machine learning baselines, we adopt default hyperparameters. Our BioBERT-based model embeds free-text chief complaints into a 768-dimensional representation using the BioBERT-mnli-snli-scinli-scitail-mednli-stsb sentence transformer. These embeddings are concatenated with vital signs and passed through a two-layer neural network (100 hidden units per layer) implemented in sklearn.

For LLM predictions, we use gpt-4o-2024-08-06 with a temperature of 0. Experiments on the MIMIC data are conducted in a HIPAA-compliant Azure OpenAI environment, ensuring adherence to the MIMIC data usage agreement.

Among our baselines, two are notable: KATE, in-context learning with KNN-based retrieval of examples, and fine-tuning on top of continued pretraining. For our implementation, KATE retrieves the $3k$ most similar training examples using BioBERT embeddings of the unstructured text e.g. chief complaints. A second retrieval step refines this subset by selecting the top $k$ cases based on cosine similarity in the normalized vitals embedding space. For fine-tuning on top of continued pretraining, we first extend the unsupervised training of Llama-3.1-8B-Instruct and Qwen3-8B using the ESI handbook for 10 epochs. To prevent data leakage, we exclude all sections that contain training or test questions. This pretraining step is skipped for the KTAS dataset due to differences in triage protocols between ESI and KTAS. Following continued pretraining, we perform supervised fine-tuning on the labelled triage datasets for 25 epochs, except for MIMIC, where we run it for 10 epochs due to its larger dataset size. All fine-tuning is conducted using LoRA \cite{hu2022lora}.

To determine the optimal number of supervised fine-tuning epochs, we evaluated 10, 25, and 100 epochs, finding that 25 epochs provided the best trade-off between performance and overfitting. Across all experiments, for continued pretraining, we use a learning rate of 4e-5, weight decay of 0.01, LoRA rank $r$ = 32, scaling factor $\alpha$ = 32, a cosine learning rate scheduler with a warm-up ratio of 0.1, and a batch size of 8. For task-specific supervised fine-tuning, we adopt a learning rate of 2e-4, weight decay of 0.01, the same LoRA parameters (r = 32, $\alpha = 32$), a linear learning rate scheduler with five warm-up steps, and a batch size of 8.

\begin{table}[h]
\tbl{Evaluation metrics on the test set for four metrics (F-1, Accuracy, QWK, MSE) across three datasets: MIMIC, KTAS, and ESI-Handbook. Higher values indicate better performance for all metrics except MSE, where lower values are preferable. The best-performing baseline for each metric is bolded, and the second-best and the third-best are underlined. We recommend KTAS as the reference dataset for resolving any discrepancies between findings, due to its large size and gold-standard expert annotations.}
{
\centering
\resizebox{\textwidth}{!}{%
\setlength{\tabcolsep}{3pt}
\begin{tabular}{@{\extracolsep{\fill}}ccccccccccccc@{}}
\hline
  & \multicolumn{4}{c}{\textsc{ESI-Handbook}} 
  & \multicolumn{4}{c}{\textsc{KTAS}} 
  & \multicolumn{4}{c}{\textsc{MIMIC}} \\
Methodology & QWK   & MSE $(\downarrow)$  & F-1   & Acc.  & QWK   & MSE $(\downarrow)$  & F-1   & Acc.  & QWK   & MSE $(\downarrow)$  & F-1   & Acc.  \\
\hline
\textsc{LogReg} \raggedright                             &  N.A.     &     N.A.  &   N.A.    &   N.A.    & 0.369 & 0.813 & 0.418 & 0.472 & 0.306 & 0.550 & 0.483 & 0.564 \\
\textsc{XGBoost} \raggedright                            &   N.A.    &  N.A.     &   N.A.    &   N.A.    & 0.059 & 1.475 & 0.291 & 0.342 & 0.216 & 1.616 & 0.211 & 0.210 \\
\textsc{BioBERT} \raggedright                            & 0.117 & 2.904 & 0.251 & 0.287 & 0.074 & 1.169 & 0.261 & 0.330 & 0.498 & 0.432 & \underline{0.640} & \underline{0.656} \\
\textsc{Vanilla} \raggedright                            & 0.923 & 0.213 & 0.788 & 0.787 & 0.599 & 0.634 & 0.511 & 0.520 & 0.562 & 0.561 & 0.585 & 0.568 \\
\textsc{AutoCoT} \raggedright                            & \underline{0.932} & \underline{0.185} & \underline{0.816} & \underline{0.815} & 0.568 & 0.668 & 0.493 & 0.506 & 0.486 & 0.665 & 0.581 & 0.566 \\
\textsc{CoT} \raggedright                                & 0.909 & 0.225 & 0.809 & 0.809 & 0.598 & 0.660 & 0.520 & 0.522 & 0.386 & 0.734 & 0.514 & 0.510 \\
\textsc{FewShot} \raggedright (40 shots)                 & \textbf{0.940} & \textbf{0.169} & \textbf{0.826} & \textbf{0.831} & 0.603 & 0.639 & 0.516 & 0.518 & 0.546 & 0.494 & 0.567 & 0.563 \\
\textsc{FewShotCoT} \raggedright (40 shots)              & 0.894 & 0.292 & 0.708 & 0.708 & 0.624 & 0.573 & 0.559 & 0.561 & 0.550 & 0.486 & 0.591 & 0.588 \\
\textsc{KATE} \raggedright (40 shots)                    & \underline{0.934} & \underline{0.180} & \underline{0.821} & \underline{0.820} & \textbf{0.667} & \textbf{0.413} & \textbf{0.696} & \textbf{0.718} & \textbf{0.654} & \textbf{0.352} & \textbf{0.677} & \textbf{0.678} \\
\textsc{KATECoT} \raggedright (40 shots)                 & 0.917 & 0.236 & 0.763 & 0.764 & \textbf{0.667} & \underline{0.497} & \underline{0.607} & \underline{0.608} & \underline{0.601} & \underline{0.402} & \underline{0.639} & \underline{0.640} \\
\textsc{Llama-8B} \raggedright (fine-tuned) & 0.724 & 0.551 & 0.538 & 0.584 & \underline{0.656} & \underline{0.511} & 0.591 & 0.589 & \underline{0.595} & \underline{0.417} & 0.626 & 0.637 \\
\textsc{Qwen-8B} \raggedright (fine-tuned) & 0.901 & 0.354 & 0.693 & 0.689 & N.A. & N.A. & N.A. & N.A. & N.A. & N.A. & N.A. & N.A. \\
\textsc{Vanilla} \raggedright (o3-mini)                  & 0.828 & 0.382 & 0.632 & 0.652 & 0.645 & 0.542 & \underline{0.606} & \underline{0.604} & 0.366 & 0.803 & 0.496 & 0.491 \\
\textsc{KATE} \raggedright (o3-mini, 5 shots)           & 0.884 & 0.292 & 0.714 & 0.708 & 0.573 & 0.656 & 0.555 & 0.554 & 0.520 & 0.474 & 0.622 & 0.624 \\
\hline

\end{tabular}}
}
\end{table}

\section{Results}

In Table 1, we present the performance of various models on the triage prediction task.

\textbf{Relevance of demonstrations matters.} Among the baselines, KATE demonstrates the strongest overall performance, significantly outperforming other methods. Interestingly, its CoT variant performs worse than expected, ranking as the second-best method. While CoT generally enhances performance, comparing vanilla prompting to their CoT or AutoCoT counterparts in Table 1, there appears to be a threshold beyond which excessive demonstrations may overload the model, negatively impacting its reasoning. We further investigate this phenomenon in our ablation study.

Notably, KATE substantially outperforms few-shot prompting, underscoring the importance of retrieving relevant examples, though the ESI-Handbook stands out as a slight exception. This highlights how there is still a need to rely on ``experience" (by drawing on past examples) despite LLMs' internal knowledge of the domain. Furthermore, models like o3-mini achieve strong performance through reasoning, but this proves insufficient on its own. Strangely, when providing o3-mini with demonstrations via KATE, we observe mixed results: performance improves but not on KTAS, which is concerning since KTAS has gold labels unlike MIMIC.

Additionally, the smaller Llama-3.1-8B-Instruct model demonstrates competitive performance. While its performance is limited on the ESI-Handbook dataset due to the scarcity of training samples, it performs well on KTAS and MIMIC, where larger training sets are available, ranking third-best in terms of QWK and MSE. Preliminary results with Qwen3-8B, which outperforms Llama on the ESI-Handbook, further underscore the potential of open-source models for domain-specific specialisation.

\begin{figure}[h]
\centerline{
\includegraphics[width=0.6\textwidth]{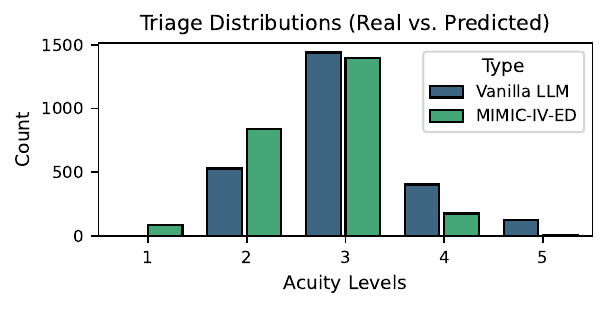} 
}
\caption{Comparison of triage distributions between MIMIC-IV-ED and GPT-4o predictions using vanilla prompting.}
\label{fig2}
\end{figure}

\textbf{MIMIC.} The MIMIC dataset is a rich resource with hundreds of thousands of samples, providing the human triage distribution at scale. One striking observation, illustrated in Figure \ref{fig2}, is the discrepancy between human-assigned triage labels in MIMIC and the predictions made by LLMs. Specifically, LLMs exhibit hesitation in assigning patients to the most severe acuity level (1), a pattern previously documented in other tasks \cite{li2024hot}. This discrepancy underscores a broader issue of calibration.

\begin{table}[h]
\tbl{Results of counterfactual analysis, testing prominent LLMs across two configurations: Vanilla Prompting and CoT. The values for each row represent the acuity assigned to patients of that category on average. Bolded values in each column indicate the most prioritized demographic group by the model, while underlined values denote the least prioritized. The row without race represents the averages across race for that particular sex. The final section (the last three rows) provides p-values for statistical significance tests, where * indicates $p < 0.05$ and ** indicates $p < 0.01$. While the rest have been corrected for multiple testing, we leave the p-values for o3-mini uncorrected to highlight its lack of bias.\label{example_table}}{
\centering
\resizebox{\textwidth}{!}{%
\setlength{\tabcolsep}{3pt}
\begin{tabular}{@{\extracolsep{\fill}}lccccccccccc}
\hline
Demographic & \multicolumn{1}{c}{\textsc{Llama}} &  \multicolumn{1}{c}{\textsc{Gemini}} & \multicolumn{2}{c}{\textsc{gpt-4o-mini}} & \multicolumn{2}{c}{\textsc{gpt-4o}} & \multicolumn{2}{c}{\textsc{claude-3-haiku}} & \multicolumn{2}{c}{\textsc{claude-3-sonnet}} & \multicolumn{1}{c}{\textsc{o3}}\\ 
                             &  \textsc{3.1-70B}       &   \textsc{2.0}      & Vanilla          & CoT            & Vanilla          & CoT            & Vanilla          & CoT                  & Vanilla          & CoT  & \textsc{mini}              \\ 
\hline
Men                       & 2.478    & 2.860    & 2.759    & 2.630    & 3.018    & 2.977    & 2.591    & 2.894    & 2.948    & 3.094    & 2.869 \\
\quad American Indian     & \textbf{2.458}    & 2.863    & 2.753    & 2.666    & \underline{3.181}    & 2.983    & \underline{2.800}    & 2.912    & 3.000    & 3.132    & 2.858 \\
\quad Asian               & 2.468    & 2.890    & 2.769    & 2.612    & 3.045    & 3.000    & 2.500    & \underline{2.947}    & \underline{3.095}    & 3.066    & 2.853 \\
\quad Black               & 2.474    & \underline{2.895}    & 2.584    & 2.573    & 2.931    & 2.923    & 2.566    & 2.883    & 2.714    & 3.080    & 2.879 \\
\quad Hispanic            & 2.495    & \textbf{2.811}    & 2.846    & 2.627    & \textbf{2.886}    & 2.940    & 2.683    & \underline{2.947}    & 2.928    & 3.073    & 2.863 \\
\quad Native Hawaiian     & \textbf{2.458}    & 2.874    & 2.553    & 2.565    & 3.022    & 3.059    & 2.216    & 2.848    & 2.857    & \underline{3.161}    & \underline{2.905} \\
\quad White               & \underline{2.516}    & 2.832    & \underline{3.046}    & \underline{2.736}    & 3.045    & 2.957    & 2.783    & 2.830    & \underline{3.095}    & 3.051    & 2.858 \\
\hline
Women                     & 2.486    & 2.865    & 2.653    & 2.629    & 3.064    & 2.963    & 2.436    & 2.810    & 2.682    & 3.068    & 2.865 \\
\quad American Indian     & 2.463    & 2.874    & 2.692    & 2.666    & \underline{3.181}    & 2.889    & 2.516    & 2.824    & 2.785    & 3.154    & 2.863 \\
\quad Asian               & 2.495    & 2.874    & 2.707    & 2.589    & 3.045    & 3.042    & 2.366    & 2.824    & 2.642    & 3.029    & 2.847 \\
\quad Black               & 2.479    & 2.858    & 2.584    & 2.627    & 3.022    & \textbf{2.881}    & 2.333    & 2.824    & \textbf{2.523}    & \textbf{3.022}    & 2.874 \\
\quad Hispanic            & 2.500    & 2.868    & 2.800    & 2.705    & 3.022    & 2.906    & 2.500    & 2.824    & 2.547    & 3.044    & 2.890 \\
\quad Native Hawaiian     & 2.474    & 2.847    & \textbf{2.384}    & \textbf{2.511}    & 3.000    & \underline{3.093}    & \textbf{2.166}    & \textbf{2.713}    & 2.595    & 3.073    & \textbf{2.837} \\
\quad White               & 2.505    & 2.868    & 2.753    & 2.674    & 3.113    & 2.966    & 2.733    & 2.853    & 3.000    & 3.088    & 2.879 \\
\hline
\quad Sex                 & 1.00     & 1.00     & $<0.01^{**}$   & 1.00     & 1.00     & 1.00     & $<0.01^{**}$   & $<0.01^{**}$   & $<0.01^{**}$   & 1.00    & 0.728 \\
\quad Race                & $<0.01^{**}$   & 1.00     & $<0.01^{**}$   & $<0.01^{**}$   & $<0.01^{**}$   & $<0.01^{**}$   & $<0.01^{**}$   & 1.00     & $<0.01^{**}$   & 0.247   & 0.824 \\
\quad Sex \& Race         & $<0.01^{**}$   & 0.39     & $<0.01^{**}$   & 0.028$^{*}$    & 0.52     & 0.01$^{*}$    & $<0.01^{**}$   & 0.056    & $<0.01^{**}$   & 0.83    & 0.693 \\
\hline

\end{tabular}}}
\end{table}

\textbf{Intersectional Biases.} Our counterfactual analysis reveals significant demographic preferences across many of the evaluated models, with consistent patterns emerging. Notably, biases are most pronounced at specific intersections. For example, Native Hawaiian women are strongly favoured by GPT and Claude models. Under GPT-4o-mini with vanilla prompting, they receive an average acuity score of 2.38 compared to 3.05 for White men (lower scores mean higher priority), reflecting nearly a full triage level difference. Claude models, in particular, show broader preference toward Native Hawaiian, Hispanic, and Black patients. Some biases are model-specific—for instance, GPT-4o favours Hispanic men, unlike its mini variant. In contrast, Asian and White patients are never the most favoured demographic, while other groups benefit from model-specific variation. In particular, white men consistently emerge as the least preferred group across all models.  

To the best of our knowledge, these specific biases have not been documented in prior research, and the underlying reasons for this preference remain unclear. Given that these biases contradict societal disparities typically observed in healthcare, we hypothesize that they may be a downstream effect of post-training alignment. 

Interestingly, CoT prompting consistently reduces the bias, suggesting that explicit reasoning enables models to overlook irrelevant demographic attributes. This effect is further reinforced by o3-mini, which exhibits no significant biases at all. Moreover, Gemini-2.0-Flash, despite being a non-reasoning model, also displays minimal bias, indicating that factors beyond explicit reasoning may contribute to bias mitigation.

\subsection{Ablations}

\textbf{Does domain-specific knowledge help foundational models?} Recent foundational models have been shown to outperform domain-specific models, especially in medicine \cite{nori2023can}. However, the low performance of these LLMs on our task suggests that their pretraining corpora may contain limited coverage of triage-specific protocols. To investigate this, we explore two ways of continued pretraining on the ESI-Handbook: additional epochs, and paraphrasing the original text—a strategy observed to improve performance without inducing overfitting\cite{allen2023physics,chang2024large}. Our paraphrasing approach involves segmenting the ESI handbook by paragraph, prompting GPT-4.1 to paraphrase each segment, and then reassembling the modified text. We define a notation of the form $n \cdot k$, where $k$ denotes the number of paraphrased handbooks used, and $n$ indicates the number of epochs over each. For example, a $6 \cdot 5$ paraphrased mixing setup corresponds to training over five paraphrased versions of the handbook, each for six epochs. We still fine-tune on the labeled dataset for 25 epochs afterwards.

Interestingly, naïve continued pretraining without paraphrasing underperforms the baseline (68.90\% vs. 70.81\% accuracy), suggesting repeated exposure of a single text isn't sufficient for learning. In contrast, paraphrased mixing yields substantial improvements, peaking at 76.56\% accuracy, before degrading with excessive epochs. These results challenge the prevailing view that effective adaptation requires vast amounts of domain-specific data. Instead, even a single document, with thoughtful augmentation, can meaningfully improve LLM performance.

\begin{table}[h]
\tbl{Evaluation of Continued Pretraining Strategies on ESI-Handbook with Qwen3-8B.\label{table_pretraining}}{
\centering
\resizebox{0.45\textwidth}{!}{%
\setlength{\tabcolsep}{6pt}
\begin{tabular}{@{\extracolsep{\fill}}lcc}
\hline
\textbf{Method} & \textbf{Accuracy} & \textbf{QWK} \\
\hline
Without continued pre-training                          & 70.81\% & 0.8814 \\
10 epochs                          & 68.90\% & 0.9006 \\
30 epochs                          & 69.38\% & 0.8809 \\
30 epochs (with 6 $\cdot$ 5 paraphrased mixing) & 73.68\% & 0.8863 \\
30 epochs (with 3 $\cdot$ 10 paraphrased mixing) & 76.56\% & 0.9058 \\
60 epochs (with 6 $\cdot$ 10 paraphrased mixing) & 63.64\% & 0.8289 \\
\hline
\end{tabular}}}
\end{table}

\textbf{Does model precision impact deployment feasibility?} As hospitals seek to deploy language models efficiently on local infrastructure, model size and memory footprint become critical considerations. We compare 4-bit and 16-bit precision versions of our models and find that 4-bit quantization offers significant reductions in memory usage, but also performance degradation, highlighting the outstanding limitations of lower-precision models. 

\begin{table}[h]
\tbl{Impact of Model Precision on Performance on ESI-Handbook with Qwen3-8B.\label{table_precision}}{
\centering
\resizebox{0.45\textwidth}{!}{%
\setlength{\tabcolsep}{6pt}
\begin{tabular}{@{\extracolsep{\fill}}lcc}
\hline
\textbf{Precision Setting} & \textbf{Accuracy} & \textbf{QWK} \\
\hline
4-bit  & 43.06\% & 0.7294 \\
16-bit & 68.90\% & 0.9006 \\
\hline
\end{tabular}}}
\end{table}

\textbf{Does serialization affect the LLM?} In the context of LLMs, serialization refers to converting tabular data into natural language. We experiment with different serialization methods for presenting triage data and find that structuring the data as a natural paragraph is not particularly important. Instead, a simple format—listing feature names sequentially, separated by columns, followed by their corresponding values in the same order—yields better performance. For all our results in Table 1, we use the natural serialization method.

\begin{table}[h]
\tbl{Evaluation of Serialization Strategies on KTAS and MIMIC.\label{table_serialization}}{
\centering
\resizebox{0.35\textwidth}{!}{%
\setlength{\tabcolsep}{3pt}
\begin{tabular}{@{\extracolsep{\fill}}lcc}
\hline
                         & \textbf{KTAS} & \textbf{MIMIC} \\
Serialization            & F-1 Score       & F-1 Score  \\
\hline
Commas   & \multicolumn{1}{c}{0.569} & 0.582 \\
Natural                  & 0.511       & 0.585 \\
Newlines                 & 0.499       & 0.580 \\
Spaces                   & 0.540       & 0.582 \\
\hline
\end{tabular}}
}

\end{table}

\begin{figure}[h]
\centerline{
\includegraphics[width=0.8\textwidth]{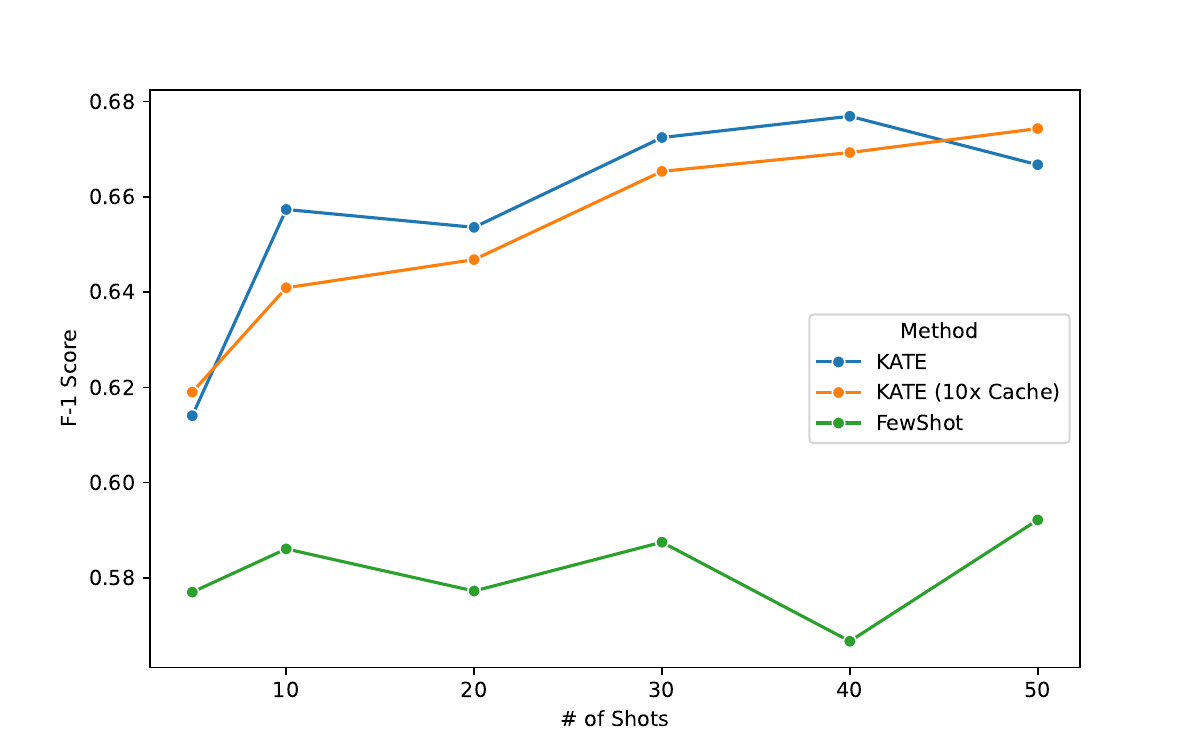} 
}
\caption{Few-shot Scaling (On MIMIC-IV)}
\label{fig3}
\end{figure}

\textbf{Do the Number of Shots Matter?} Given the importance of examples, as demonstrated by KATE’s effectiveness, we investigate how performance scales with the number of shots. Our results in Figure \ref{fig3} reveal key nuances: Few-shot prompting struggles to improve with additional samples, whereas KATE exhibits consistent, monotonic gains. Furthermore, expanding KATE’s retrieval pool—by increasing its accessible training set tenfold—has minimal direct impact but enhances its scaling behavior.

\textbf{Do the Number of Shots Affect Chain-of-Thought?} To examine the interaction between demonstration count and chain-of-thought reasoning, we conduct an ablation on KTAS, where gold labels are more reliable than in MIMIC. KATE’s performance increases at 40 shots but drops at 50, indicating that the gains from increasing shots are not monotonic, consistent with our prior ablation results. When incorporating chain-of-thought prompting (KATECoT), we observe a more nuanced trend. While accuracy metrics show variability, Cohen’s Kappa—a more robust measure of agreement—reveals a consistent improvement with additional demonstrations.

\begin{figure}[h]
\centerline{
\includegraphics[width=0.8\textwidth]{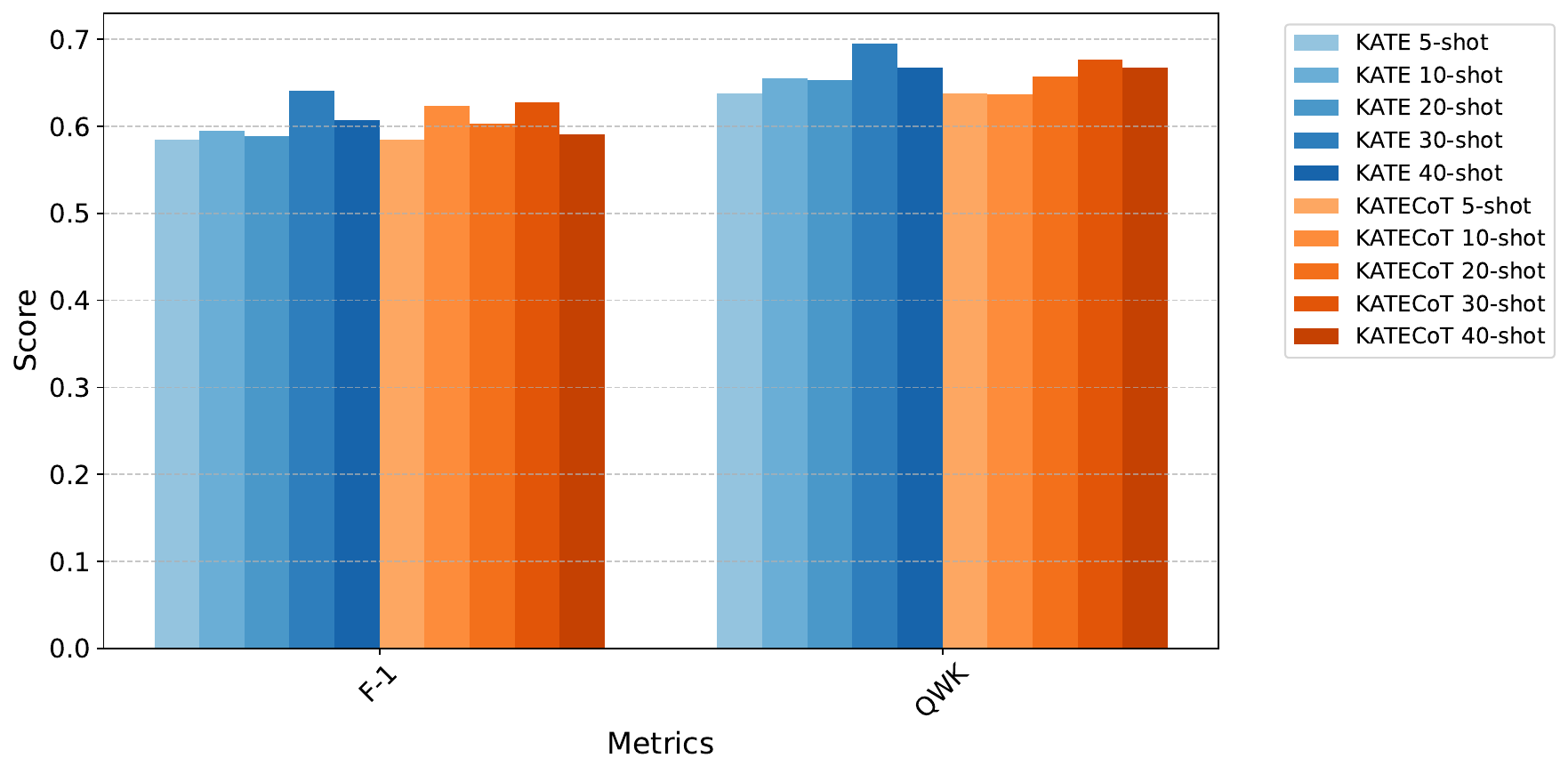} 
}
\caption{Comparison of KATE and KATECoT Performance Across Different Shot Counts (On KTAS)}
\label{fig4}
\end{figure}

\section{Discussion}

While our findings highlight the promising capabilities of LLMs and outline a potential future for their integration into emergency room workflows, we acknowledge that this study represents only an early-stage, preclinical evaluation. It does not constitute a formal clinical trial and should be viewed as a stage 1 assessment within the broader clinical validation pipeline. Furthermore, our findings are subject to limitations. A more exhaustive exploration of alternative configurations—such as retrieval-augmented generation (RAG) or full fine-tuning, diverse machine learning architectures, and a wider range of fine-tuning hyperparameters—could yield further insights and improved performance. Notably, dataset quality plays a crucial role in our experiments: unlike KTAS and the ESI Handbook, where labels are based on expert consensus or curated ground truth, MIMIC reflects real-time triage decisions made by nurses in clinical settings, which may include errors or inconsistencies and lack retrospective validation. This may explain why reasoning-based methods like CoT and AutoCoT show reduced performance on MIMIC. These observations underscore the need for high-quality, large-scale gold-standard datasets to advance triage modelling.

Our findings also highlight the emergence of demographic disparities in LLM predictions. Our aim is not to prescribe normative conclusions about which groups should or should not be prioritised. Rather, these findings highlight the importance of transparent auditing frameworks to surface unexpected behaviours in model predictions—especially in high-stakes settings like clinical triage. We leave the ethical and medical implications of such disparities to domain experts and clinicians, and emphasise that these tools should ultimately serve to support fair and accountable decision-making.

Beyond healthcare, we find that this work offers an investigative framework for assessing the deployment of AI in similar rankings-based, high-stakes decision-making contexts, such as recruitment, college admissions, and academic evaluations.

\section{Conclusion}

This study explores the opportunities and challenges of utilizing LLMs for triage in EDs. Our findings demonstrate the significant potential of LLMs, which substantially outperform traditional machine learning baselines. Notably, retrieving relevant examples enables the LLM to make more informed decisions, exhibiting logic that closely resembles that of clinicians who draw on past experiences with patients. However, key challenges remain, particularly in model calibration (e.g., hesitance to assign the highest acuity level) and bias. Our study reveals that many of the prominent models have encoded demographic preferences, favoring certain groups over others. These biases raise significant ethical concerns, particularly in clinical triage, where prioritization has a direct impact on patient outcomes. Moreover, integration into clinical workflows requires more than strong performance or minimal bias. The objective is not to replace nurses, but to support them in making triage more efficient, equitable, and accurate. Realising this vision will require design choices that prioritise usability, interpretability, and seamless alignment with existing clinical practices. While a full exploration of clinical integration is beyond the scope of this work, we hope our study serves as a foundational step—by addressing two critical pillars of clinical AI, predictive accuracy and bias—and provides a springboard for future research that brings these systems closer to safe, real-world deployment.





\section*{Acknowledgments}
This work was supported in part by NIH grants P30 AG073105, U01 AG066833, and U01 AG068057.

\bibliographystyle{ws-procs11x85}
\bibliography{aaai25}

\begin{thebibliography}{10}

\bibitem{cascella2023evaluating}
M.~Cascella, J.~Montomoli, V.~Bellini and E.~Bignami, Evaluating the feasibility of chatgpt in healthcare: an analysis of multiple clinical and research scenarios, {\em Journal of medical systems} {\bf 47}, p.~33  (2023).

\bibitem{park2024assessing}
Y.-J. Park, A.~Pillai, J.~Deng, E.~Guo, M.~Gupta, M.~Paget and C.~Naugler, Assessing the research landscape and clinical utility of large language models: A scoping review, {\em BMC Medical Informatics and Decision Making} {\bf 24}, p.~72  (2024).

\bibitem{li2024scoping}
L.~Li, J.~Zhou, Z.~Gao, W.~Hua, L.~Fan, H.~Yu, L.~Hagen, Y.~Zhang, T.~L. Assimes, L.~Hemphill {\em et~al.}, A scoping review of using large language models (llms) to investigate electronic health records (ehrs), {\em arXiv preprint arXiv:2405.03066}   (2024).

\bibitem{jiang2023health}
L.~Y. Jiang, X.~C. Liu, N.~P. Nejatian, M.~Nasir-Moin, D.~Wang, A.~Abidin, K.~Eaton, H.~A. Riina, I.~Laufer, P.~Punjabi {\em et~al.}, Health system-scale language models are all-purpose prediction engines, {\em Nature} {\bf 619}, 357  (2023).

\bibitem{liu2023using}
S.~Liu, A.~P. Wright, B.~L. Patterson, J.~P. Wanderer, R.~W. Turer, S.~D. Nelson, A.~B. McCoy, D.~F. Sittig and A.~Wright, Using ai-generated suggestions from chatgpt to optimize clinical decision support, {\em Journal of the American Medical Informatics Association} {\bf 30}, 1237  (2023).

\bibitem{montagna2023data}
S.~Montagna, S.~Ferretti, L.~C. Klopfenstein, A.~Florio and M.~F. Pengo, Data decentralisation of llm-based chatbot systems in chronic disease self-management, in {\em Proceedings of the 2023 ACM Conference on Information Technology for Social Good\/},  2023.

\bibitem{franc2024accuracy}
J.~M. Franc, A.~J. Hertelendy, L.~Cheng, R.~Hata and M.~Verde, Accuracy of a commercial large language model (chatgpt) to perform disaster triage of simulated patients using the simple triage and rapid treatment (start) protocol: Gage repeatability and reproducibility study, {\em Journal of Medical Internet Research} {\bf 26}, p. e55648  (2024).

\bibitem{chmielewski2022esi}
N.~Chmielewski and J.~Moretz, Esi triage distribution in us emergency departments, {\em Advanced emergency nursing journal} {\bf 44}, 46  (2022).

\bibitem{friedman2024artificial}
A.~B. Friedman, M.~K. Delgado and G.~E. Weissman, Artificial intelligence for emergency care triage—much promise, but still much to learn, {\em JAMA Network Open} {\bf 7}, e248857  (2024).

\bibitem{hinson2018accuracy}
J.~S. Hinson, D.~A. Martinez, P.~S. Schmitz, M.~Toerper, D.~Radu, J.~Scheulen, S.~A. Stewart~de Ramirez and S.~Levin, Accuracy of emergency department triage using the emergency severity index and independent predictors of under-triage and over-triage in brazil: a retrospective cohort analysis, {\em International journal of emergency medicine} {\bf 11}, 1  (2018).

\bibitem{joseph2023race}
J.~W. Joseph, M.~Kennedy, A.~M. Landry, R.~H. Marsh, E.~B. Da’Marcus, D.~E. Im, P.~C. Chen, M.~E. Samuels-Kalow, L.~M. Nentwich, N.~Elhadad {\em et~al.}, Race and ethnicity and primary language in emergency department triage, {\em JAMA Network Open} {\bf 6}, e2337557  (2023).

\bibitem{hong2018predicting}
W.~S. Hong, A.~D. Haimovich and R.~A. Taylor, Predicting hospital admission at emergency department triage using machine learning, {\em PloS one} {\bf 13}, p. e0201016  (2018).

\bibitem{goto2019machine}
T.~Goto, C.~A. Camargo, M.~K. Faridi, R.~J. Freishtat and K.~Hasegawa, Machine learning--based prediction of clinical outcomes for children during emergency department triage, {\em JAMA network open} {\bf 2}, e186937  (2019).

\bibitem{miles2020using}
J.~Miles, J.~Turner, R.~Jacques, J.~Williams and S.~Mason, Using machine-learning risk prediction models to triage the acuity of undifferentiated patients entering the emergency care system: a systematic review, {\em Diagnostic and prognostic research} {\bf 4}, 1  (2020).

\bibitem{singhal2023large}
K.~Singhal, S.~Azizi, T.~Tu, S.~S. Mahdavi, J.~Wei, H.~W. Chung, N.~Scales, A.~Tanwani, H.~Cole-Lewis, S.~Pfohl {\em et~al.}, Large language models encode clinical knowledge, {\em Nature} {\bf 620}, 172  (2023).

\bibitem{goh2023chatgpt}
E.~Goh, B.~Bunning, E.~Khoong, R.~Gallo, A.~Milstein, D.~Centola and J.~H. Chen, Chatgpt influence on medical decision-making, bias, and equity: a randomized study of clinicians evaluating clinical vignettes, {\em Medrxiv}   (2023).

\bibitem{preiksaitis2024role}
C.~Preiksaitis, N.~Ashenburg, G.~Bunney, A.~Chu, R.~Kabeer, F.~Riley, R.~Ribeira and C.~Rose, The role of large language models in transforming emergency medicine: Scoping review, {\em JMIR Medical Informatics} {\bf 12}, p. e53787  (2024).

\bibitem{johns2022insight}
{Johns Hopkins Medicine}, Insight nov./dec. 2022, {\em Insight}  (November 2022).

\bibitem{williams2024use}
C.~Y. Williams, T.~Zack, B.~Y. Miao, M.~Sushil, M.~Wang, A.~E. Kornblith and A.~J. Butte, Use of a large language model to assess clinical acuity of adults in the emergency department, {\em JAMA Network Open} {\bf 7}, e248895  (2024).

\bibitem{masanneck2024triage}
L.~Masanneck, L.~Schmidt, A.~Seifert, T.~K{\"o}lsche, N.~Huntemann, R.~Jansen, M.~Mehsin, M.~Bernhard, S.~G. Meuth, L.~B{\"o}hm {\em et~al.}, Triage performance across large language models, chatgpt, and untrained doctors in emergency medicine: Comparative study, {\em Journal of Medical Internet Research} {\bf 26}, p. e53297  (2024).

\bibitem{lu2024triageagent}
M.~Lu, B.~Ho, D.~Ren and X.~Wang, Triageagent: Towards better multi-agents collaborations for large language model-based clinical triage, in {\em Findings of the Association for Computational Linguistics: EMNLP 2024\/},  2024.

\bibitem{guevara2024large}
M.~Guevara, S.~Chen, S.~Thomas, T.~L. Chaunzwa, I.~Franco, B.~H. Kann, S.~Moningi, J.~M. Qian, M.~Goldstein, S.~Harper {\em et~al.}, Large language models to identify social determinants of health in electronic health records, {\em NPJ digital medicine} {\bf 7}, p.~6  (2024).

\bibitem{omiye2023large}
J.~A. Omiye, J.~C. Lester, S.~Spichak, V.~Rotemberg and R.~Daneshjou, Large language models propagate race-based medicine, {\em NPJ Digital Medicine} {\bf 6}, p. 195  (2023).

\bibitem{zack2024assessing}
T.~Zack, E.~Lehman, M.~Suzgun, J.~A. Rodriguez, L.~A. Celi, J.~Gichoya, D.~Jurafsky, P.~Szolovits, D.~W. Bates, R.-E.~E. Abdulnour {\em et~al.}, Assessing the potential of gpt-4 to perpetuate racial and gender biases in health care: a model evaluation study, {\em The Lancet Digital Health} {\bf 6}, e12  (2024).

\bibitem{yang2024unmasking}
Y.~Yang, X.~Liu, Q.~Jin, F.~Huang and Z.~Lu, Unmasking and quantifying racial bias of large language models in medical report generation, {\em ArXiv}   (2024).

\bibitem{yunxiang2023chatdoctor}
L.~Yunxiang, L.~Zihan, Z.~Kai, D.~Ruilong and Z.~You, Chatdoctor: A medical chat model fine-tuned on llama model using medical domain knowledge, {\em arXiv preprint arXiv:2303.14070}   (2023).

\bibitem{shu2023visual}
C.~Shu, B.~Chen, F.~Liu, Z.~Fu, E.~Shareghi and N.~Collier, Visual med-alpaca: A parameter-efficient biomedical llm with visual capabilities  (2023).

\bibitem{chen2023meditron}
Z.~Chen, A.~H. Cano, A.~Romanou, A.~Bonnet, K.~Matoba, F.~Salvi, M.~Pagliardini, S.~Fan, A.~K{\"o}pf, A.~Mohtashami {\em et~al.}, Meditron-70b: Scaling medical pretraining for large language models, {\em arXiv preprint arXiv:2311.16079}   (2023).

\bibitem{zhang2023biomedgpt}
K.~Zhang, J.~Yu, Z.~Yan, Y.~Liu, E.~Adhikarla, S.~Fu, X.~Chen, C.~Chen, Y.~Zhou, X.~Li {\em et~al.}, Biomedgpt: A unified and generalist biomedical generative pre-trained transformer for vision, language, and multimodal tasks, {\em arXiv preprint arXiv:2305.17100}   (2023).

\bibitem{labrak2024biomistral}
Y.~Labrak, A.~Bazoge, E.~Morin, P.-A. Gourraud, M.~Rouvier and R.~Dufour, Biomistral: A collection of open-source pretrained large language models for medical domains, {\em arXiv preprint arXiv:2402.10373}   (2024).

\bibitem{jeong2024improving}
M.~Jeong, J.~Sohn, M.~Sung and J.~Kang, Improving medical reasoning through retrieval and self-reflection with retrieval-augmented large language models, {\em arXiv preprint arXiv:2401.15269}   (2024).

\bibitem{toufiq2023harnessing}
M.~Toufiq, D.~Rinchai, E.~Bettacchioli, B.~S.~A. Kabeer, T.~Khan, B.~Subba, O.~White, M.~Yurieva, J.~George, N.~Jourde-Chiche {\em et~al.}, Harnessing large language models (llms) for candidate gene prioritization and selection, {\em Journal of Translational Medicine} {\bf 21}, p. 728  (2023).

\bibitem{lee2024knowledgedrivenfeatureselectionengineering}
J.~Lee, S.~Yang, J.~Y. Baik, X.~Liu, Z.~Tan, D.~Li, Z.~Wen, B.~Hou, D.~Duong-Tran, T.~Chen and L.~Shen, Knowledge-driven feature selection and engineering for genotype data with large language models  (2024).

\bibitem{li-etal-2024-dalk}
D.~Li, S.~Yang, Z.~Tan, J.~Y. Baik, S.~Yun, J.~Lee, A.~Chacko, B.~Hou, D.~Duong-Tran, Y.~Ding, H.~Liu, L.~Shen and T.~Chen, {DALK}: Dynamic co-augmentation of {LLM}s and {KG} to answer {A}lzheimer{'}s disease questions with scientific literature, in {\em Findings of the Association for Computational Linguistics: EMNLP 2024\/},  eds. Y.~Al-Onaizan, M.~Bansal and Y.-N. Chen (Association for Computational Linguistics, Miami, Florida, USA, November 2024).

\bibitem{zakka2024almanac}
C.~Zakka, R.~Shad, A.~Chaurasia, A.~R. Dalal, J.~L. Kim, M.~Moor, R.~Fong, C.~Phillips, K.~Alexander, E.~Ashley {\em et~al.}, Almanac—retrieval-augmented language models for clinical medicine, {\em NEJM AI} {\bf 1}, p. AIoa2300068  (2024).

\bibitem{lozano2023clinfo}
A.~Lozano, S.~L. Fleming, C.-C. Chiang and N.~Shah, Clinfo. ai: An open-source retrieval-augmented large language model system for answering medical questions using scientific literature, in {\em PACIFIC SYMPOSIUM ON BIOCOMPUTING 2024\/},  2023.

\bibitem{shang2024leveragingsocialdeterminantshealth}
T.~Shang, S.~Yang, W.~He, T.~Zhai, D.~Li, B.~Hou, T.~Chen, J.~H. Moore, M.~D. Ritchie and L.~Shen, Leveraging social determinants of health in alzheimer's research using llm-augmented literature mining and knowledge graphs  (2024).

\bibitem{rubin2021learning}
O.~Rubin, J.~Herzig and J.~Berant, Learning to retrieve prompts for in-context learning, {\em arXiv preprint arXiv:2112.08633}   (2021).

\bibitem{xu2023knn}
B.~Xu, Q.~Wang, Z.~Mao, Y.~Lyu, Q.~She and Y.~Zhang, $ k $ nn prompting: Beyond-context learning with calibration-free nearest neighbor inference, in {\em The Eleventh International Conference on Learning Representations\/},  2023.

\bibitem{ovadia2023fine}
O.~Ovadia, M.~Brief, M.~Mishaeli and O.~Elisha, Fine-tuning or retrieval? comparing knowledge injection in llms, {\em arXiv preprint arXiv:2312.05934}   (2023).

\bibitem{zhou2023lima}
C.~Zhou, P.~Liu, P.~Xu, S.~Iyer, J.~Sun, Y.~Mao, X.~Ma, A.~Efrat, P.~Yu, L.~Yu {\em et~al.}, Lima: Less is more for alignment, {\em Advances in Neural Information Processing Systems} {\bf 36}, 55006  (2023).

\bibitem{ito2023accuracy}
N.~Ito, S.~Kadomatsu, M.~Fujisawa, K.~Fukaguchi, R.~Ishizawa, N.~Kanda, D.~Kasugai, M.~Nakajima, T.~Goto and Y.~Tsugawa, The accuracy and potential racial and ethnic biases of gpt-4 in the diagnosis and triage of health conditions: evaluation study, {\em JMIR Medical Education} {\bf 9}, p. e47532  (2023).

\bibitem{williams2024evaluating}
C.~Y. Williams, B.~Y. Miao, A.~E. Kornblith and A.~J. Butte, Evaluating the use of large language models to provide clinical recommendations in the emergency department, {\em Nature Communications} {\bf 15}, p. 8236  (2024).

\bibitem{li2024mediq}
S.~S. Li, V.~Balachandran, S.~Feng, J.~S. Ilgen, E.~Pierson, P.~W. Koh and Y.~Tsvetkov, Mediq: Question-asking llms and a benchmark for reliable interactive clinical reasoning, {\em arXiv preprint arXiv:2406.00922}   (2024).

\bibitem{ouyang-etal-2024-climedbench}
Z.~Ouyang, Y.~Qiu, L.~Wang, G.~De~Melo, Y.~Zhang, Y.~Wang and L.~He, {C}li{M}ed{B}ench: A large-scale {C}hinese benchmark for evaluating medical large language models in clinical scenarios, in {\em Proceedings of the 2024 Conference on Empirical Methods in Natural Language Processing\/},  eds. Y.~Al-Onaizan, M.~Bansal and Y.-N. Chen (Association for Computational Linguistics, Miami, Florida, USA, November 2024).

\bibitem{hofmann2024ai}
V.~Hofmann, P.~R. Kalluri, D.~Jurafsky and S.~King, Ai generates covertly racist decisions about people based on their dialect, {\em Nature} {\bf 633}, 147  (2024).

\bibitem{Johnson2023MIMICIVED}
A.~Johnson, L.~Bulgarelli, T.~Pollard, L.~A. Celi, R.~Mark and S.~Horng, {MIMIC-IV-ED (version 2.2)}  (2023).

\bibitem{moon2019triage}
S.-H. Moon, J.~L. Shim, K.-S. Park and C.-S. Park, Triage accuracy and causes of mistriage using the korean triage and acuity scale, {\em PloS one} {\bf 14}, p. e0216972  (2019).

\bibitem{Gilboy_2005}
N.~Gilboy, {\em The emergency severity index. version 4: Implementation handbook} (U.S. Dept. of Health and Human Services, Public Health Service, Agency for Healthcare Research and Quality, 2005).

\bibitem{lee2020biobert}
J.~Lee, W.~Yoon, S.~Kim, D.~Kim, S.~Kim, C.~H. So and J.~Kang, Biobert: a pre-trained biomedical language representation model for biomedical text mining, {\em Bioinformatics} {\bf 36}, 1234  (2020).

\bibitem{wei2022chain}
J.~Wei, X.~Wang, D.~Schuurmans, M.~Bosma, F.~Xia, E.~Chi, Q.~V. Le, D.~Zhou {\em et~al.}, Chain-of-thought prompting elicits reasoning in large language models, {\em Advances in Neural Information Processing Systems} {\bf 35}, 24824  (2022).

\bibitem{zhang2022automatic}
Z.~Zhang, A.~Zhang, M.~Li and A.~Smola, Automatic chain of thought prompting in large language models, {\em arXiv preprint arXiv:2210.03493}   (2022).

\bibitem{liu-etal-2022-makes}
J.~Liu, D.~Shen, Y.~Zhang, B.~Dolan, L.~Carin and W.~Chen, What makes good in-context examples for {GPT}-3?, in {\em Proceedings of Deep Learning Inside Out (DeeLIO 2022): The 3rd Workshop on Knowledge Extraction and Integration for Deep Learning Architectures\/},  eds. E.~Agirre, M.~Apidianaki and I.~Vuli{\'c} (Association for Computational Linguistics, Dublin, Ireland and Online, May 2022).

\bibitem{hu2022lora}
E.~J. Hu, Y.~Shen, P.~Wallis, Z.~Allen-Zhu, Y.~Li, S.~Wang, L.~Wang, W.~Chen {\em et~al.}, Lora: Low-rank adaptation of large language models., {\em ICLR} {\bf 1}, p.~3  (2022).

\bibitem{li2024hot}
L.~Li, L.~Fan, S.~Atreja and L.~Hemphill, “hot” chatgpt: The promise of chatgpt in detecting and discriminating hateful, offensive, and toxic comments on social media, {\em ACM Transactions on the Web} {\bf 18}, 1  (2024).

\bibitem{nori2023can}
H.~Nori, Y.~T. Lee, S.~Zhang, D.~Carignan, R.~Edgar, N.~Fusi, N.~King, J.~Larson, Y.~Li, W.~Liu {\em et~al.}, Can generalist foundation models outcompete special-purpose tuning? case study in medicine, {\em arXiv preprint arXiv:2311.16452}   (2023).

\bibitem{allen2023physics}
Z.~Allen-Zhu and Y.~Li, Physics of language models: Part 3.1, knowledge storage and extraction, {\em arXiv preprint arXiv:2309.14316}   (2023).

\bibitem{chang2024large}
H.~Chang, J.~Park, S.~Ye, S.~Yang, Y.~Seo, D.-S. Chang and M.~Seo, How do large language models acquire factual knowledge during pretraining?, {\em Advances in neural information processing systems} {\bf 37}, 60626  (2024).

\end{thebibliography}

\end{document}